\pdfoutput=1
\documentclass[a4paper]{article}

\usepackage{INTERSPEECH2018}
\def \paragraph {\vskip0.5\normalbaselineskip \noindent \textbf}
\usepackage{multirow}
\usepackage{amssymb}
\usepackage{amsmath}
\usepackage[usenames, dvipsnames]{color}
\usepackage{graphicx}
\usepackage{subfigure}
\usepackage{booktabs}

\title{Multimodal Polynomial Fusion for Detecting Driver Distraction}
\name{Yulun Du, Chirag Raman, Alan W Black, Louis-Philippe Morency, Maxine Eskenazi}
\address{
  Language Technologies Institute, Carnegie Mellon University, USA
}
\email{\{yulund, craman, awb, morency, max\}@cs.cmu.edu}

\begin{document}

\maketitle
\begin{abstract}
Distracted driving is deadly, claiming 3,477 lives in the U.S. in 2015 alone. 
Although there has been a considerable amount of research
on modeling the distracted behavior of drivers under various conditions, accurate automatic detection using multiple modalities and especially the contribution of using the speech modality to improve accuracy has received little attention.
This paper introduces a new multimodal dataset for distracted driving behavior and discusses automatic distraction detection using features from three modalities: facial expression, speech and car signals.
Detailed multimodal feature analysis shows that adding more modalities monotonically increases the predictive accuracy of the model.
Finally, a simple and effective multimodal fusion technique using a polynomial fusion layer shows superior distraction detection results compared to the baseline SVM and neural network models.

\end{abstract}

\section{Introduction}
In 2015, 3477 deaths in car crashes in the U.S. were attributed to distracted driving \cite{nhtsa}. The use of electronic devices, particularly cellphones, while driving is one of the most common causes. 
Vehicle and cellphone manufacturers designed speech interfaces (such as Siri) that were supposed to reduce potential distraction by eliminating the need to look at a screen.
However, studies show that hands-free voice technologies are still highly distracting \cite{handsfree}. While legislation in many states bans the use of cellphones while driving, many individuals continue to speak and text while driving. Since this practice continues, another strategy must be adopted: warning the driver when a dangerous situation arises while she is distracted. 

Automatic distraction detection can enable in-car systems or virtual personal assistants to choose the right time to warn the driver,  giving out safety information, or shut down some app in a dangerous situation.
Early attempts to do this had high false alarm rates \cite{fpr}. False alarms cause drivers to ignore or disable the system. This lack of robustness is often linked to the paucity of information, that is, the use of only one or two modalities for detection, often just facial expressions. 
Detecting driver distraction is complex.
A variety of modalities come into play, all of which should be used to detect distraction. In addition to facial expression, there is the driver's speech while talking to a passenger, instructing an intelligent agent or talking on the phone. Information coming from the vehicle itself (CAN bus) such as knowing when the driver is braking is also important. Interaction across modalities is important. For example, a system based on eye-gaze alone fails if the driver is wearing sunglasses.
Recent advances in multimodal deep learning afford better performance thanks to both improved facial feature detection and the ability to learn a joint representation from multiple modalities via multimodal fusion \cite{mmml}. There are two advantages of multimodal deep learning over traditional unimodal approaches: 1) better accuracy, and 2) more robust detection. 

We have developed a novel deep multimodal polynomial fusion (MPF) architecture to robustly detect distraction. Specifically, we used a polynomial function to map features from different modalities to a weighted sum of the intermodal product interactions as the fused representation for distraction detection. We also introduce a new training and assessment dataset. 


The contribution of this paper is three-fold:
\begin{itemize}
\item A database of distracted driving behavior containing distraction events.
\item Empirical evidence that incorporating multiple modalities improves distraction detection.
\item A simple and effective multimodal fusion technique that outperforms baseline models.
\end{itemize}

\section{Related Work}
This section describes existing distraction detection approaches for each modality, and then describes previous work which explored multiple modalities. We also present related datasets.

\paragraph{Visual - Facial expression}
Previous work focused on facial cues such as facial landmarks, head pose turns, glances, eye-gaze tracking, and facial action units \cite{yulan-svm, glance_lex, fernandez2016driver, dl-face, au, kang2013various}.
While good detection accuracy was achieved without the use of other modalities, these approaches lack robustness in cases where either some part of the face is obscured or the lighting changes dramatically \cite{fernandez2016driver}. Data from additional modalities can compensate for the missing information.

\paragraph{Visual - Road conditions}
Researchers used a forward-facing camera and computer vision algorithms \cite{lane, road, cvpr-eye-road, cmu-eye-road}. 
Scene understanding is used along with information from a backward-facing camera (driver's glances) to categorize driving behavior \cite{cvpr-eye-road,cmu-eye-road}. This bi-modal approach can detect what the driver is attending to on the road ahead \cite{utd2}. 
Lane position changes can be captured by the forward-facing camera \cite{lane,cmu-eye-road}.

\paragraph{Acoustics - Speech}
The driver's speech has been used by \cite{utd, speech, Craye2016}. There are voice interfaces installed in the vehicle, such as a spoken dialog system or personal assistant \cite{utd, utd2}. The driver's speech is analyzed to derive features such as voice activity detection and strings of words via automatic speech recognition \cite{Craye2016}.

\paragraph{Driving measures}
Vehicle control signals also encode changes in driving performance that reflect distraction \cite{Craye2016}. They serve as complementary information to other modalities. CAN-Bus information that has been used includes: speed, steering wheel position, gas pedal usage, and break pedal usage \cite{utd, utd2, kang2013various, lane}. 

\paragraph{Combinations of modalities} There have been attempts to use multimodal fusion for distraction detection \cite{utd, utd2, Craye2016, cvpr-eye-road, multi-li}. These used early fusion techniques that concatenate multimodal features into a single feature vector.
For example, \cite{multi-li} uses modalities similar to those used in this paper. They used data from front and rear-facing cameras to extract road and facial features, audio from a microphone to extract the energy of speech, and CAN-bus information. They performed early feature fusion, trained machine learning models for binary classification of distraction, and showed promising results. This did not, however, model the intermodal interaction amongst features \cite{tfn}. We compare our MPF model to their best binary classification model, SVM, below.

\paragraph{Other information}
A few studies have used body position from a Kinect camera \cite{kinect}. This indicates whether the driver's both hands are on the steering wheel.
Physiological signals such as an electroencephalogram (EEG) have also been studied \cite{kang2013various}. 
Both approaches require additional equipment not commonly found on vehicles (often due to cost). Due to the limited amount of existing data and the fact that these signals are noisy, they are less interesting for distraction detection.

\paragraph{Datasets}
One of the most well-studied distraction detection datasets is UTDrive \cite{utd}. Collected in the 2000s, it is naturalistic and multimodal, and has a speech interface. 
Beside the limited sensor capabilities existing at the time the data was recorded, the dataset does not have: 1) extensive dialog interaction (i.e. phone usage), and 2) sufficient amounts of data. 
Taamneh et al. \cite{newest-data} released a multimodal distracted driving dataset which is large enough for our needs, but does not include recorded speech. To investigate distraction detection using multimodal deep learning, we need a dataset that has speech and more instances of distraction. 



\section{Multimodal Distraction Dataset}

We designed a dataset that can capture more instances and nuances of distraction (we have $\sim147k$ datapoints at frame-level in the training set and $\sim25\%$ of them correspond to places where the driver was distracted). Specifically, we wanted to create distracting instances that afford different degrees of cognitive load, due to the road conditions, the type of message to be dealt with and the coincidence of the two. We also wanted to represent several sources of messages: texts, phone calls and emails. We recorded as many different modalities as possible (forward camera, backward camera, microphone, car information). There were 30 subjects, 7 female and 23 male, each driving for about 15 minutes (minimum 9 minutes, maximum 21.36 minutes). 

While driving, each subject had three types of interaction at different levels of cognitive load with the message agent on an Android phone. For example, low cognitive load is a combination of low-load email from Mom asking if the driver was feeling ok today while the subject was driving on a straightaway. High load is the combination of a friend asking the driver to list five things she wants for her upcoming birthday while she is entering a hairpin turn.  Messages were sent as: text messages, phone calls, and email. The four modalities we captured were time synchronized. After driving, the subject was asked to watch the recording and annotate stretches of time (start and end point) when they felt they had been distracted.

\paragraph{Data collection system architecture}
The simulated driving route was created using the OpenDS driving simulator \cite{opends}. Traffic signs, a traffic light, hairpin turns, and odd objects along the side of the road created situations that demanded  attention. The simulator recorded and synchronized the driving data.
A spoken dialog system was connected to a personal assistant that interacted with the driver to produce the predefined messages.
The MultiSense recorder \footnote{http://multicomp.cs.cmu.edu/} recorded and synchronized facial videos and speech from a backward-facing camera.
Open Broadcaster screen capture software \footnote{https://obsproject.com/}, served as the forward-facing camera, capturing what the subject saw on the screen while driving.
A wizard interface controlled all of the submodules and initiated tasks in the dialog system.
Each type of signal was synchronized with timestamps and saved to the database. 


\begin{figure}[t]
  \centering
  \includegraphics[width=1.0\linewidth]{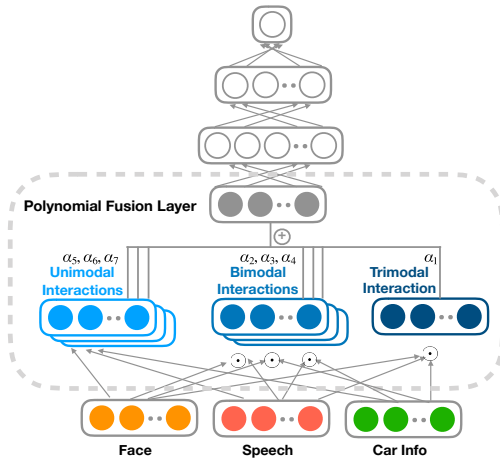}
  \caption{Proposed Multimodal Polynomial Fusion model}
  \label{fig:model}
\end{figure}

The multimodal distraction detection dataset will be made publicly available after Interspeech publication of this paper.

\section{Multimodal Polynomial Fusion}
In this section, we present multimodal polynomial fusion (MPF) for learning fused representations for distraction detection.

Given feature vectors $x_{F}$, $x_{S}$, $x_{C}$ at each time frame (10Hz) from face, speech, and car modalities respectively, we want to learn a shared hidden representation $h_{fusion}$ that captures the interaction amongst these modalities to detect driver distraction at that time frame.

The cube activation function $f_{cube}(z) = (z)^3$ \cite{chen}, where $z = h_1 + h_2 + ... + h_n + \beta_0$, is a simple way to learn such a shared representation from multiple feature vectors ($h_1, h_2, ..., h_n$, along with a bias term $\beta_0$) of the same dimension. 
It is a special configuration of Polynomial Networks \cite{poly1, poly2, poly3,qizhe1,qizhe2}.
Chen et al. \cite{chen} show that the product interactions of features captured by cube activation empirically lead to a better representation for dependency parsing.  
Intuitively, the cube activation function resembles a polynomial kernel that extracts 3-combinations with repetitions from $h_{1}$, $h_{2}$, ..., $h_{n}$, and $\beta_0$:
\begin{equation}
 \begin{split}
   (h_{1} + h_{2} + ... + h_{n} + \beta_0)^3 = \sum_{i, j, k \in \{1, ..., n\}} h_{i} \odot h_{j} \odot h_{k} \\ 
   + \beta_0 \odot \sum_{i, j \in \{1, ..., n\}} h_{i} \odot h_{j} + \beta_0^2 \odot \sum_{i \in \{1, ..., n\}} h_{i} + \beta_0^3\\
 \end{split}
\end{equation}
where $\odot$ is the Hadamard product.

The multimodal polynomial fusion (MPF) layer is inspired by the cube activation function. First, the feature vectors from each modality are transformed so that all of them have the same dimension $|h|$, so that an element-wise product (the Hadamard product) of features from different modalities can be performed:
\begin{equation}
 \begin{split}
  h_{F} = W_{F} x_{F},\quad h_{S} = W_{S} x_{S},\quad h_{C} = W_{C} x_{C} \\
  \label{eq0}
 \end{split}
\end{equation}
where $W_{F} \in \mathbb{R}^{|h|\times |x_{F}|}$, $W_{S} \in \mathbb{R}^{|h|\times |x_{S}|}$, $W_{C} \in \mathbb{R}^{|h|\times |x_{C}|}$. Then $h_{F}$, $h_{S}$, and $h_{C}$ could be passed to a cube activation to model the intermodal interactions of the three: $h_{fusion} = (h_{F} + h_{S} + h_{C} + \beta_0)^3$.

However, cube activation captures redundant combinations from duplicated feature modalities, although being computationally efficient.
For example, $h_{F} \odot h_{F} \odot h_{C}$ is also an interaction term captured by cube activation, but it is not a reasonable representation to include in multimodal fusion, since modalities need not be repeatedly used in intermodal interaction. Such redundancy could increase complexity and lead to inferior predictive results.

To alleviate this problem, the multimodal polynomial fusion (MPF) layer (Eq.~\ref{eq1}) is designed to model $h_{fusion}$ by summing up selected weighted intermodal interaction amongst the three feature modalities. Each interaction is derived by an element-wise product of features. The proposed neural architecture is shown on Figure~\ref{fig:model}. The multimodal fusion representation is calculated using the following polynomial function:
\begin{equation}
 \begin{split}
  h_{MPF} = f_{MPF}(h_{F}, h_{S}, h_{C}) = (\alpha_0 \cdot h_{F} \odot h_{S} \odot h_{C} + \\
  \alpha_1 \cdot h_{F} \odot h_{S} + \alpha_2 \cdot h_{F} \odot h_{C} + \alpha_3 \cdot h_{S} \odot h_{C} \\
  + \alpha_4 \cdot h_{F} + \alpha_5 \cdot h_{S} + \alpha_6 \cdot h_{C} + \beta_0)
  \label{eq1}
 \end{split}
\end{equation}
where $\alpha_i \in \mathbb{R}$ are learnable parameters adjusting the weight of each term, $\beta_0 \in \mathbb{R}^{|h|}$ is the bias term, and $\odot$ is the element-wise multiplication of vectors. Thus, $h_{F} \odot h_{S} \odot h_{C}$ models trimodal interaction; $h_{F} \odot h_{C}$, $h_{F} \odot h_{S}$, and $h_{S} \odot h_{C}$ models bimodal interaction; $h_{F}$, $h_{S}$, and $h_{C}$ are the unimodal features. Essentially, $h_{MPF}$ is a weighted sum of the product interactions of the feature vectors from the three modalities.

The advantage of the polynomial fusion layer is that it explicitly specifies the desired combinations of modalities to model interaction and also learns the weights for all intermodal dynamics. The polynomial fusion layer can easily be extended to accommodate more modalities by adding more terms in the polynomial.

We then feed the fused hidden representation $h_{MPF}$ to a tanh activation: 
\begin{equation}
  h_{fusion} = tanh(h_{MPF})
  \label{eq2}
\end{equation}
The $tanh$ activation function is used because $h_{MPF}$ is unbounded and thus needs to be controlled by bounded non-linear activation \cite{pei}. Empirically, we found that $tanh$ did stabilize network training and led to better results than unbounded activations such as $ReLU$.

Finally, the hidden representation $h_{fusion}$ is fed to a two-layer feed forward neural network with dropouts and ReLU activations. The complete model is shown on Figure~\ref{fig:model}.

\begin{figure}[t]
  \centering
  \includegraphics[width=\linewidth]{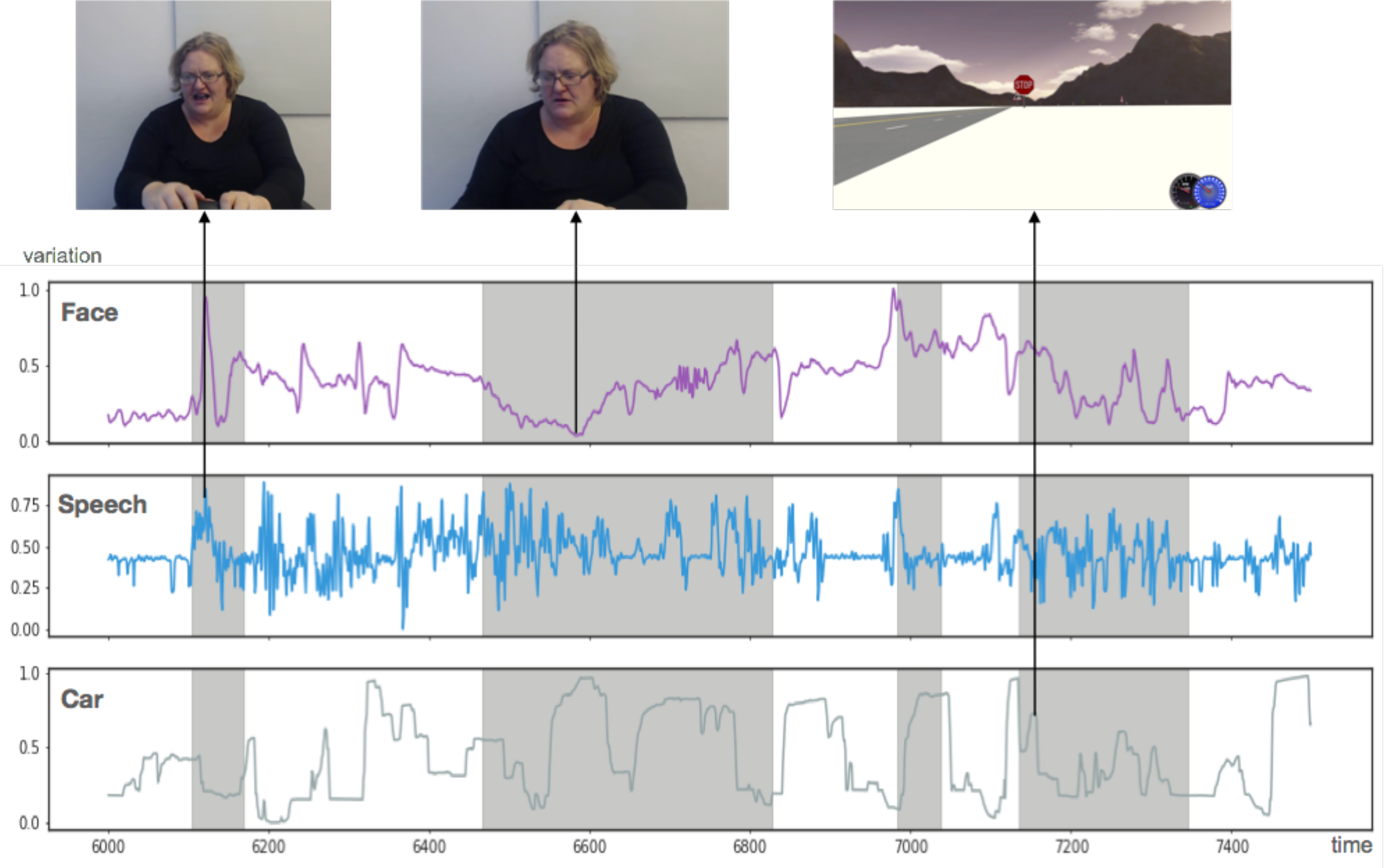}
  \caption{A case-in-point visualization of distraction for three modalities: Face, Speech, and Car. Each feature modality is reduced and normalized to a one-dimensional space and projected onto a continuous time axis. The grey area denotes the time period when the driver said she was distracted.}
  \label{fig:eg}
\end{figure}

\section{Experiments}
We now describe the multimodal features, baseline models, experimental methodology, and results.

\subsection{Multimodal Features}
This paper focuses on three modalities: facial expression/movement, speech, and car information. Feature sets were extracted from the backward facing camera video and the speech signal. 


\paragraph{Facial features} The OpenFace \cite{openface} toolkit is a state-of-the-art tool used for facial landmark detection, head pose estimation, facial action unit recognition, and eye-gaze estimation. The facial features we use include:
a) \textit{Facial landmarks (FL):} 68 points on the face (204 values).
b) \textit{Gaze vectors:} 3D vector and gaze angle for each eye (8 values), and 28 2D and 3D eye region landmarks for each eye (280 values).
c) \textit{18 Action Units (AU):} regression and binary outputs of all the available AUs in \cite{openface} (36 values).
d) \textit{Head pose:} 3D translation and 3D rotation of head pose (6 numbers).
Facial features are extracted at a frame rate of 30Hz.

\paragraph{Speech features} The OpenSMILE \cite{opensmile} toolkit is an audio feature extractor that extracts a knowledge-based feature set. The speech-related features include:
a) \textit{Prosody:} Pitch and loudness.
b) \textit{Voice-Quality (VQ):} jitter and shimmer, creaky voice.
c) \textit{Frame Energy}.
d) \textit{Voice Activity Detection (VAD)}.
e) \textit{F0 fundamental frequency}.
f) \textit{Syllables per second (SPS)}.
The speech-related features are extracted with a moving window of 300ms and a shift of 100ms.

\paragraph{Car driving measures} The features from the car logged by the driving simulator are:
a) \textit{Speed of the vehicle:} a real number (in km/h).
b) \textit{Steering wheel position:} a continuous number from -1 to 1.
c) \textit{Gas pedal position:} a continuous number from 0 to 1.
d) \textit{Break pedal position:} a continuous number from 0 to 1.

All of the above features were synchronized with respect to the frame of audio features (10Hz). Distraction classification is performed here on a frame-wise basis.

\paragraph{Qualitative feature analysis}
Feature value variation by dimensionality reduction for each of the three modalities over time is shown for one example in Figure~\ref{fig:eg}. The grey area denotes the time period when the driver said she was distracted. Figure~\ref{fig:eg} shows that when the driver screamed, there was a peak in the speech features as well as a corresponding peak in the facial features, due to mouth opening. Similarly, when the driver turned her head to the side toward the phone, the facial features show a negative peak. There was also a significant change in car information showing the moment when the driver realized that she went off the road and tried to shift back onto it. This qualitative analysis shows that the feature peaks may correlate with and complement one another in indicating driver distraction (the grey area). Thus, distraction detection may benefit from modeling the intermodal interaction of the modalities.

\subsection{Baseline Models}
We compare our model (\textit{MPF}) to seven baseline models and \textit{MPF} variants: \textit{Majority} is the trivial baseline predicting the majority label; \textit{SVM} is the Support Vector Machine using early fusion multimodal features in \cite{multi-li} which achieves the best performance on binary classification of distraction; \textit{NN-Early} is a two-layer feed forward neural network that takes the concatenation of features from three modalities; \textit{NN-Cube} is the cube activation function for fusing features from three modalities described in \cite{chen}; \textit{NN-TC} is the tanh-cube activation function for fusing features from three modalities described in \cite{pei}; \textit{MPF-1} and  \textit{MPF-2}, which are variants of our full \textit{MPF} model, use one modality and two modalities as input respectively. To ensure fair comparisons of neural network models, all of the fused representations have the same size (except for early fusion which has a larger size due to concatenation) and are fed to a two-layer feed forward neural network with the same number of parameters using the train/dev/test partition mentioned below.


  

\begin{table}[th]
  \caption{Distraction detection results on the multimodal distraction dataset test portion. Our model outperforms the baseline models for unweighted accuracy (Acc), Area Under Curve (AUC), Equal Error Rate (EER), and F-1 score. (For EER only, the lower the score the better the performance.)} 
  \label{tab:ret}
  \centering
  
\begin{tabular}{llllll}  
\toprule
Model    & Modal. & Acc. & AUC & EER & F-1\\
\midrule
Majority    & --    & 0.7753  &  0.5000 & 0.5000 & -- \\
\midrule
          & F    & 0.7749  & 0.6752 & 0.3714 & 0.4965  \\
MPF-1     & S     & 0.7491  & 0.5271 & 0.4850 & 0.1816   \\
          & C     & 0.7724   & 0.5141 & 0.4928 & 0.0813    \\
\midrule
          & F + S    & 0.7976  & 0.6960 & 0.3568 & 0.5318   \\
MPF-2     & F + C     & 0.7932   & 0.6935 & 0.3579 & 0.5269   \\
          & S + C     & 0.7633   & 0.5386 & 0.4787 & 0.1987   \\
\midrule
SVM     & All     & 0.7542   & 0.6637 & 0.3768 & 0.4772   \\
NN-Early     & All   & 0.8046   & 0.6867 & 0.3693 & 0.5208    \\
NN-Cube    & All   & 0.8023   & 0.7048 & 0.3488 & 0.5453  \\
NN-TC    & All   & 0.8015   & 0.6931 & 0.3615 & 0.5290  \\
\midrule
MPF    & All    & \textbf{0.8139}   & \textbf{0.7152} & \textbf{0.3416} & \textbf{0.5641}   \\
\bottomrule
\end{tabular}

\end{table}



\subsection{Methodology}
The 30 subjects were randomly separated into 20/5/5 train/dev/test sets. Each subject is in only one partition so that models can generalize to new drivers. For each subject, features are scaled to zero mean and unit variance. The binary classification of distraction is performed at frame-level, so the train/dev/test sets have 147k/36k/37k datapoints. Since the data is imbalanced, the performance of the models is evaluated by Area Under ROC Curve (AUC), Equal Error Rate (EER), and F-1 score. We chose the hyper-parameters of each model based on its development set performance. Neural network models were trained using the Adam optimizer \cite{adam} with a step learning rate scheduler, regularized by dropouts \cite{drop}. The size of the fused representation $|h|$ is 16 and the size of hidden layers is 8. 

\subsection{Results and Discussion}
In Table~\ref{tab:ret}, we report the experimental results of the baseline models and our \textit{MPF} model with accuracy, AUC, EER, and F-1 score. For \textit{MPF-1} and \textit{MPF-2}, we also show the choice of feature combinations that was used. Other baseline models besides \textit{Majority} use all three modalities. 

Table~\ref{tab:ret} shows that \textit{MPF} performs best for the combinations of modalities (\textit{MPF-1} and \textit{MPF-2}). It also shows that even if some unimodal features have poor performance, all modalities contribute to a certain extent to the results. That is, we  achieve monotonically increasing accuracy with MPF using modalities that may not individually have good performance. This emphasizes the importance of intermodal interaction in multimodal fusion representation.

Results show that MPF performs better than the baseline models (all using three modalities), where the \textit{MPF} model achieves an AUC of 0.7152, an EER of 0.3416, and an F-1 score of 0.5641 on the test set, while the best baseline \textit{NN-Cube} achieves 0.7048, 0.3488, and 0.5453 respectively.

\begin{figure}[t]
  \centering
  \includegraphics[width=0.6\linewidth]{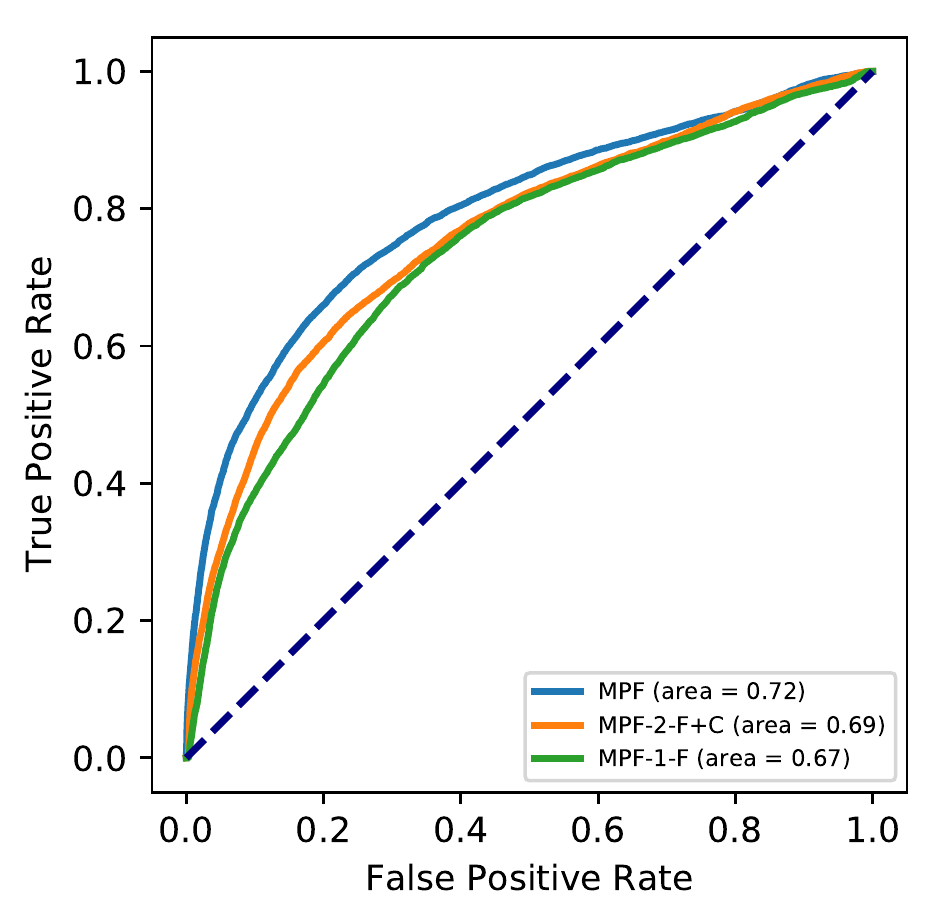}
  \caption{ROC curve of MPF (all three modalities), MPF-2 (facial and car modalities), and MPF-1 (facial modality).}
  \label{fig:roc}
\end{figure}

We also show the ROC curve of selected models in Figure~\ref{fig:roc}.  at various detection thresholds. We see that using more modalities (blue curve) has the largest ROC AUC. Performance increased when the speech modality was added. By adjusting the detection threshold, \textit{MPF} achieves the lowest false positive rate while preserving good detection accuracy.


\section{Conclusion}
The results confirm that combining signals from multiple modalities through MPF affords better prediction performance for distraction detection due to its ability to model unimodal, bimodal and trimodal interactions. 
In future work we plan to add the fourth modality from the forward-facing camera that records road conditions to further boost performance.

\section{Acknowledgements}
We thank the anonymous reviewers for their valuable comments. We thank Zhenqiang Xu and Qizhe Xie for suggestions on the draft.
This work has been sponsored by the U.S. Department of Transportation grant (Carnegie Mellon University UTC T-SET). The opinions expressed in this paper do not necessarily reflect those of the U.S. Department of Transportation.

\bibliographystyle{IEEEtran}

\bibliography{mybib}

\begin{thebibliography}{10}
\providecommand{\url}[1]{#1}
\csname url@samestyle\endcsname
\providecommand{\newblock}{\relax}
\providecommand{\bibinfo}[2]{#2}
\providecommand{\BIBentrySTDinterwordspacing}{\spaceskip=0pt\relax}
\providecommand{\BIBentryALTinterwordstretchfactor}{4}
\providecommand{\BIBentryALTinterwordspacing}{\spaceskip=\fontdimen2\font plus
\BIBentryALTinterwordstretchfactor\fontdimen3\font minus
  \fontdimen4\font\relax}
\providecommand{\BIBforeignlanguage}[2]{{%
\expandafter\ifx\csname l@#1\endcsname\relax
\typeout{** WARNING: IEEEtran.bst: No hyphenation pattern has been}%
\typeout{** loaded for the language `#1'. Using the pattern for}%
\typeout{** the default language instead.}%
\else
\language=\csname l@#1\endcsname
\fi
#2}}
\providecommand{\BIBdecl}{\relax}
\BIBdecl

\bibitem{nhtsa}
\BIBentryALTinterwordspacing
``Distracted driving 2015,'' 2017. [Online]. Available:
  \url{https://crashstats.nhtsa.dot.gov/Api/Public/ViewPublication/812381}
\BIBentrySTDinterwordspacing

\bibitem{handsfree}
D.~L. Strayer, J.~Turrill, J.~M. Cooper, J.~R. Coleman, N.~Medeiros-Ward, and
  F.~Biondi, ``Assessing cognitive distraction in the automobile,'' \emph{Human
  Factors}, vol.~57, no.~8, pp. 1300--1324, 2015.

\bibitem{fpr}
\BIBentryALTinterwordspacing
``Distraction detection algorithm evaluation,'' 2013. [Online]. Available:
  \url{https://www.nhtsa.gov/sites/nhtsa.dot.gov/files/811548.pdf}
\BIBentrySTDinterwordspacing

\bibitem{mmml}
T.~Baltrusaitis, C.~Ahuja, and L.~Morency, ``Multimodal machine learning: {A}
  survey and taxonomy,'' \emph{CoRR}, vol. abs/1705.09406, 2017.

\bibitem{yulan-svm}
Y.~Liang, M.~L. Reyes, and J.~D. Lee, ``Real-time detection of driver cognitive
  distraction using support vector machines,'' \emph{IEEE Transactions on
  Intelligent Transportation Systems}, vol.~8, no.~2, pp. 340--350, June 2007.

\bibitem{glance_lex}
L.~Fridman, H.~Toyoda, S.~Seaman, B.~Seppelt, L.~Angell, J.~Lee, B.~Mehler, and
  B.~Reimer, ``What can be predicted from six seconds of driver glances?''
  \emph{CoRR}, vol. abs/1611.08754, 2016.

\bibitem{fernandez2016driver}
A.~Fernández, R.~Usamentiaga, J.~Carús, and R.~Casado, ``Driver distraction
  using visual-based sensors and algorithms,'' \emph{Sensors}, vol.~16, no.~11,
  pp. 1--44, 2016, jCR: 2.676 - Q1 [2016].

\bibitem{dl-face}
C.~Streiffer, R.~Raghavendra, T.~Benson, and M.~Srivatsa, ``Darnet: a deep
  learning solution for distracted driving detection,'' in \emph{Middleware
  Industry}.\hskip 1em plus 0.5em minus 0.4em\relax {ACM}, 2017, pp. 22--28.

\bibitem{au}
N.~Li and C.~Busso, ``Analysis of facial features of drivers under cognitive
  and visual distractions,'' in \emph{2013 IEEE International Conference on
  Multimedia and Expo (ICME)}, July 2013, pp. 1--6.

\bibitem{kang2013various}
H.~B. Kang, ``Various approaches for driver and driving behavior monitoring: A
  review,'' in \emph{2013 IEEE International Conference on Computer Vision
  Workshops}, Dec 2013, pp. 616--623.

\bibitem{lane}
M.~Kutila, M.~Jokela, G.~Markkula, and M.~R. Rue, ``Driver distraction
  detection with a camera vision system,'' in \emph{2007 IEEE International
  Conference on Image Processing}, vol.~6, Sept 2007, pp. VI -- 201--VI -- 204.

\bibitem{road}
R.~Klette, ``Vision-based driver assistance systems,'' 2015.

\bibitem{cvpr-eye-road}
M.~Rezaei and R.~Klette, ``Look at the driver, look at the road: No
  distraction! no accident!'' in \emph{2014 IEEE Conference on Computer Vision
  and Pattern Recognition}, June 2014, pp. 129--136.

\bibitem{cmu-eye-road}
C.-W. You, N.~D. Lane, F.~Chen, R.~Wang, Z.~Chen, T.~J. Bao, M.~Montes-de Oca,
  Y.~Cheng, M.~Lin, L.~Torresani, and A.~T. Campbell, ``Carsafe app: Alerting
  drowsy and distracted drivers using dual cameras on smartphones,'' in
  \emph{Proceeding of the 11th Annual International Conference on Mobile
  Systems, Applications, and Services}, ser. MobiSys '13.\hskip 1em plus 0.5em
  minus 0.4em\relax New York, NY, USA: ACM, 2013, pp. 13--26.

\bibitem{utd2}
J.~H.~L. Hansen, C.~Busso, Y.~Zheng, and A.~Sathyanarayana, ``Driver modeling
  for detection and assessment of driver distraction: Examples from the utdrive
  test bed,'' \emph{IEEE Signal Processing Magazine}, vol.~34, no.~4, pp.
  130--142, July 2017.

\bibitem{utd}
P.~Angkititrakul, D.~Kwak, S.~Choi, J.~Kim, A.~Phucphan, A.~Sathyanarayana, and
  J.~H.~L. Hansen, ``Getting start with utdrive: Driver-behavior modeling and
  assessment of distraction for in-vehicle speech systems,'' 2007.

\bibitem{speech}
N.~Kamaruddin and A.~Wahab, ``Driver behavior analysis through speech emotion
  understanding,'' in \emph{2010 IEEE Intelligent Vehicles Symposium}, June
  2010, pp. 238--243.

\bibitem{Craye2016}
C.~Craye, A.~Rashwan, M.~S. Kamel, and F.~Karray,
  ``\BIBforeignlanguage{English}{A multi-modal driver fatigue and distraction
  assessment system},'' \emph{\BIBforeignlanguage{English}{International
  Journal of Intelligent Transportation Systems Research}}, vol.~14, no.~3, pp.
  173--194, 2016.

\bibitem{multi-li}
\BIBentryALTinterwordspacing
N.~Li and C.~Busso, ``Predicting perceived visual and cognitive distractions of
  drivers with multimodal features,'' \emph{{IEEE} Trans. Intelligent
  Transportation Systems}, vol.~16, no.~1, pp. 51--65, 2015. [Online].
  Available: \url{https://doi.org/10.1109/TITS.2014.2324414}
\BIBentrySTDinterwordspacing

\bibitem{tfn}
A.~Zadeh, M.~Chen, S.~Poria, E.~Cambria, and L.~Morency, ``Tensor fusion
  network for multimodal sentiment analysis,'' in \emph{{EMNLP}}.\hskip 1em
  plus 0.5em minus 0.4em\relax Association for Computational Linguistics, 2017,
  pp. 1103--1114.

\bibitem{kinect}
C.~Craye and F.~Karray, ``Driver distraction detection and recognition using
  {RGB-D} sensor,'' \emph{CoRR}, vol. abs/1502.00250, 2015.

\bibitem{newest-data}
S.~Taamneh, P.~Tsiamyrtzis, M.~Dcosta, P.~Buddharaju, A.~Khatri, M.~Manser,
  T.~K. Ferris, R.~C. Wunderlich, and I.~T. Pavlidis, ``A multimodal dataset
  for various forms of distracted driving.'' \emph{Scientific data}, vol.~4, p.
  170110, 2017.

\bibitem{opends}
R.~M. et~al., ``Opends: A new open-source driving simulator for research,''
  2013.

\bibitem{chen}
D.~Chen and C.~D. Manning, ``A fast and accurate dependency parser using neural
  networks.'' in \emph{EMNLP}, A.~Moschitti, B.~Pang, and W.~Daelemans,
  Eds.\hskip 1em plus 0.5em minus 0.4em\relax ACL, 2014, pp. 740--750.

\bibitem{poly1}
R.~Livni, S.~Shalev{-}Shwartz, and O.~Shamir, ``A provably efficient algorithm
  for training deep networks,'' \emph{CoRR}, vol. abs/1304.7045, 2013.

\bibitem{poly2}
R.~Livni, S.~Shalev-Shwartz, and O.~Shamir, ``On the computational efficiency
  of training neural networks,'' in \emph{Advances in Neural Information
  Processing Systems 27}, Z.~Ghahramani, M.~Welling, C.~Cortes, N.~D. Lawrence,
  and K.~Q. Weinberger, Eds.\hskip 1em plus 0.5em minus 0.4em\relax Curran
  Associates, Inc., 2014, pp. 855--863.

\bibitem{poly3}
M.~Blondel, M.~Ishihata, A.~Fujino, and N.~Ueda, ``Polynomial networks and
  factorization machines: New insights and efficient training algorithms,'' in
  \emph{{ICML}}, ser. {JMLR} Workshop and Conference Proceedings,
  vol.~48.\hskip 1em plus 0.5em minus 0.4em\relax JMLR.org, 2016, pp. 850--858.

\bibitem{qizhe1}
\BIBentryALTinterwordspacing
Q.~Xie, K.~Sun, S.~Zhu, L.~Chen, and K.~Yu, ``Recurrent polynomial network for
  dialogue state tracking with mismatched semantic parsers,'' in
  \emph{Proceedings of the 16th Annual Meeting of the Special Interest Group on
  Discourse and Dialogue}.\hskip 1em plus 0.5em minus 0.4em\relax Prague, Czech
  Republic: Association for Computational Linguistics, September 2015, pp.
  295--304. [Online]. Available: \url{http://aclweb.org/anthology/W15-4641}
\BIBentrySTDinterwordspacing

\bibitem{qizhe2}
\BIBentryALTinterwordspacing
K.~Sun, Q.~Xie, and K.~Yu, ``Recurrent polynomial network for dialogue state
  tracking,'' \emph{CoRR}, vol. abs/1507.03934, 2015. [Online]. Available:
  \url{http://arxiv.org/abs/1507.03934}
\BIBentrySTDinterwordspacing

\bibitem{pei}
W.~Pei, T.~Ge, and B.~Chang, ``An effective neural network model for
  graph-based dependency parsing.'' in \emph{ACL (1)}.\hskip 1em plus 0.5em
  minus 0.4em\relax The Association for Computer Linguistics, 2015, pp.
  313--322.

\bibitem{openface}
T.~Baltrusaitis, P.~Robinson, and L.-P. Morency, ``Openface: An open source
  facial behavior analysis toolkit.'' in \emph{WACV}.\hskip 1em plus 0.5em
  minus 0.4em\relax IEEE Computer Society, 2016, pp. 1--10.

\bibitem{opensmile}
F.~Eyben, M.~Wöllmer, and B.~Schuller, ``Opensmile: the munich versatile and
  fast open-source audio feature extractor.'' in \emph{ACM Multimedia}, A.~D.
  Bimbo, S.-F. Chang, and A.~W.~M. Smeulders, Eds.\hskip 1em plus 0.5em minus
  0.4em\relax ACM, 2010, pp. 1459--1462.

\bibitem{adam}
\BIBentryALTinterwordspacing
D.~P. Kingma and J.~Ba, ``Adam: {A} method for stochastic optimization,''
  \emph{CoRR}, vol. abs/1412.6980, 2014. [Online]. Available:
  \url{http://arxiv.org/abs/1412.6980}
\BIBentrySTDinterwordspacing

\bibitem{drop}
\BIBentryALTinterwordspacing
N.~Srivastava, G.~Hinton, A.~Krizhevsky, I.~Sutskever, and R.~Salakhutdinov,
  ``Dropout: A simple way to prevent neural networks from overfitting,''
  \emph{J. Mach. Learn. Res.}, vol.~15, no.~1, pp. 1929--1958, Jan. 2014.
  [Online]. Available: \url{http://dl.acm.org/citation.cfm?id=2627435.2670313}
\BIBentrySTDinterwordspacing

\end{thebibliography}


\end{document}